\documentclass{article}

\usepackage{arxiv}

\usepackage[utf8]{inputenc} 
\usepackage[T1]{fontenc}    
\usepackage{hyperref}       
\usepackage{url}            
\usepackage{booktabs}       
\usepackage{amsfonts}       
\usepackage{nicefrac}       
\usepackage{microtype}      
\usepackage{graphicx}
\usepackage[numbers]{natbib}
\usepackage{doi}
\graphicspath{ {./Immagini/} }
\usepackage[font={small,it}]{caption}

\title{Short-term forecast of EV charging stations occupancy probability using Big Data streaming analysis}

\author{\hspace{1mm}Francesca Soldan, Enea Bionda, Giuseppe Mauri, Silvia Celaschi \\
	Ricerca sul Sistema Energetico - RSE S.p.A.\\
	Milano, Italy \\
	\texttt{francesca.soldan@rse-web.it} \\
}

\renewcommand{\shorttitle}{ }

\hypersetup{
pdftitle={Short-term forecast of EV charging stations occupancy probability using Big Data streaming analysis},
pdfsubject={q-bio.NC, q-bio.QM},
pdfauthor={Francesca~Soldan, Enea~Bionda, Giuseppe~Mauri, Silvia~Celaschi},
pdfkeywords={electric charging infrastructure, electric vehicle, big data streaming architecture, occupancy status forecast, streaming logistic regression},
}

\begin{document}
\maketitle
\setlength{\parindent}{1em}

\begin{abstract}
The widespread diffusion of electric mobility requires a contextual expansion of the charging infrastructure. An extended collection and processing of information regarding charging of electric vehicles may turn each electric vehicle charging station into a valuable source of streaming data. Charging point operators may profit from all these data for optimizing their operation and planning activities. In such a scenario, big data and machine learning techniques would allow valorizing real-time data coming from electric vehicle charging stations. \\
This paper presents an architecture able to deal with data streams from a charging infrastructure, with the final aim to forecast electric charging station availability after a set amount of minutes from present time. Both batch data regarding past charges and real-time data streams are used to train a streaming logistic regression model, to take into account recurrent past situations and unexpected actual events. The streaming model performs better than a model trained only using historical data.\\
The results highlight the importance of constantly updating the predictive model parameters in order to adapt to changing conditions and always provide accurate forecasts.   
\end{abstract}


\section{Introduction}
\indent The Integrated National Plan for Energy and Climate (PNIEC)\cite{pniec}, published in January 2020 by the Italian Ministry of Economic Development, previews an intensive spread of electric mobility by 2030. The aim of reaching 4 millions battery electric vehicles (EVs) and 2 millions hybrid plug-in vehicles, starting from a number of 70,000 circulating EVs, is really challenging. The increase of EVs requires a contextual expansion of the charging station network \cite{arera}. The actual number of 16,700 charging points is expected to grow to 98,000-130,000, under the scenarios reported by Motus-E in a report published in 2020 \cite{motuse}. The research centre Ricerca sul Sistema Energetico (RSE S.p.A.) confirms this scenario, estimating a number of 31,500 fast charging points and 78,600 slow charging points by 2030, for a total of around 110,000 public charging points \cite{rse}. \\\indent In this context, a more efficient electric charge data processing and collection will be necessary. Each public charging station is indeed a potential data source and the exploitation and valorization of all these data can be useful for both charging point operators and final users. Charging point operators could benefit for their planning activity, while final users could receive updated information and forecasts of future occupancy status of the charging stations. \\
\indent Real-time data streams regarding charging station occupancies may be sent to a central system, allowing their integration and processing with big data and machine learning techniques. Possible final objectives could be the identification of the most appropriate collocation of new charging stations, the development of smart charging algorithms, the evaluation of the capacity of power distribution systems to handle extra charging loads and the assessment of the market value for the services provided by electric vehicles, as vehicle-to-grid solutions \cite{smartcity}. In addition, the management of data coming from electric vehicles and their charging stations has a crucial role for operation and planning future Smart Grids \cite{smartcity}. \\
\indent This paper proposes a big data streaming architecture for providing short-term forecasts of charging station occupancy probabilities. The predictive machine learning algorithm takes into account both recurrent situations linked to the past and actual unexpected events, in order to forecast the occupancy status more accurately. The importance of considering a mix of historical and actual conditions has been stressed during the Covid-19 disease 2019 pandemic: it has introduced a multitude of disruptions to daily life, which conventional forecasting models can not correctly predict. As regards the electric energy sector, this problem arises in the context of electrical consumption forecasts on the distribution grid \cite{covid}. However, mobility restrictions during lock-downs have also impacted charging habits of EV owners, with inevitable influences on predictions of occupancies of electric charging stations.

\section{Data and methods}
In order to retrieve the occupancy probability of an EV charging station a classification model can be exploited. The logistic regression model is one of the most fundamental and widely used classification methods \cite{logregr} and has been selected for a first architectural development. \\
\indent However, a forecast model trained just using historical data can result in large forecasting errors, especially in the case of unexpected events. A prime example could be related to charging stations close to a stadium or an exhibition center: a model trained only with historical data can provide high occupancy probabilities just over days with yearly recurring events, while the occupancy probability will be low in other cases. \\
\indent As a consequence, the idea has been to initialize a Logistic Regression model with historical data related to past charges and to increasingly update the model using real-time data from the actual occupancy of EV charging stations. In this way both recurrent situations linked to the past and actual unexpected events can be taken into account. \\
\indent The conversion of available data about past charges into continuous data streams has allowed the development and testing of a big data streaming architecture, potentially able to manage real-time data coming from EV charging stations.

\subsection{Available data}
\begin{figure}[htbp]
	\centering
	\includegraphics[width=13cm]{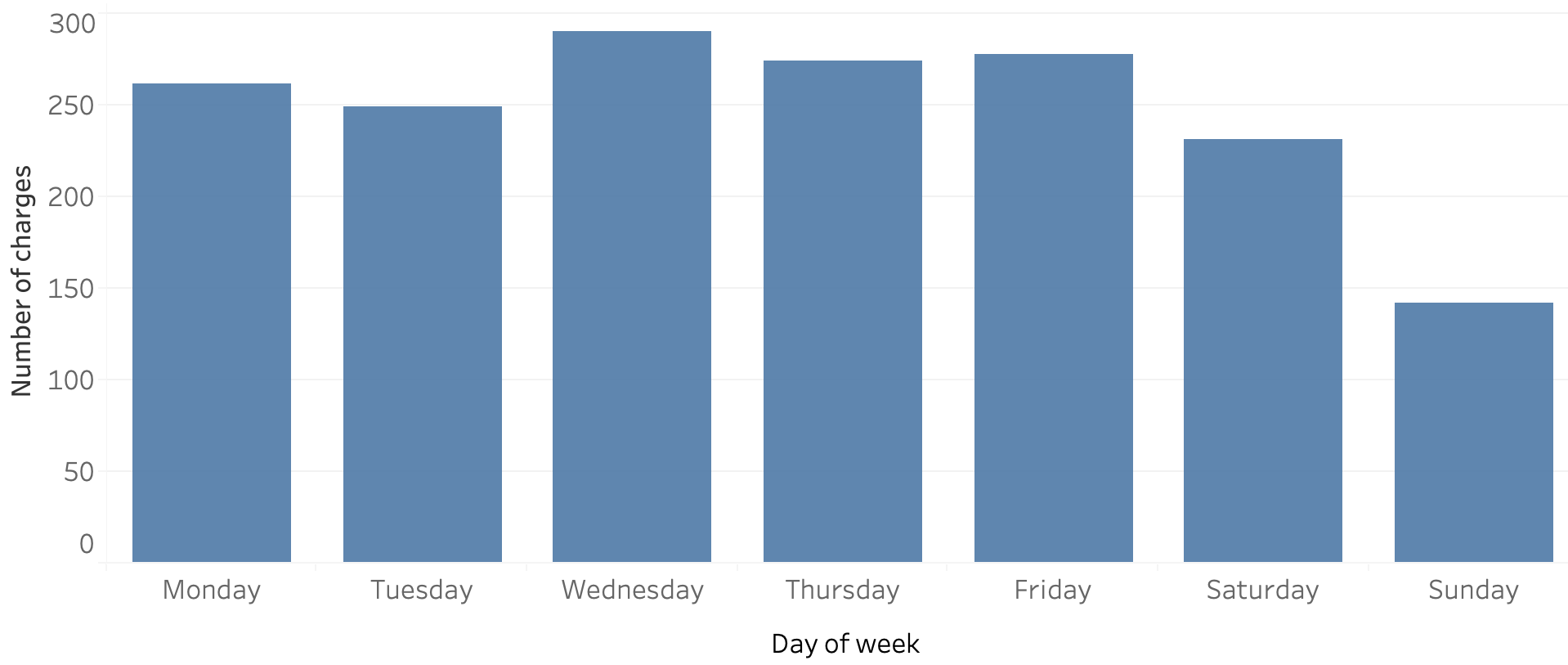}
	\caption{Weekly distribution of charges.}
	\label{Fig4}
\end{figure}

\begin{figure}[htbp]
	\centering
	\includegraphics[width=14cm]{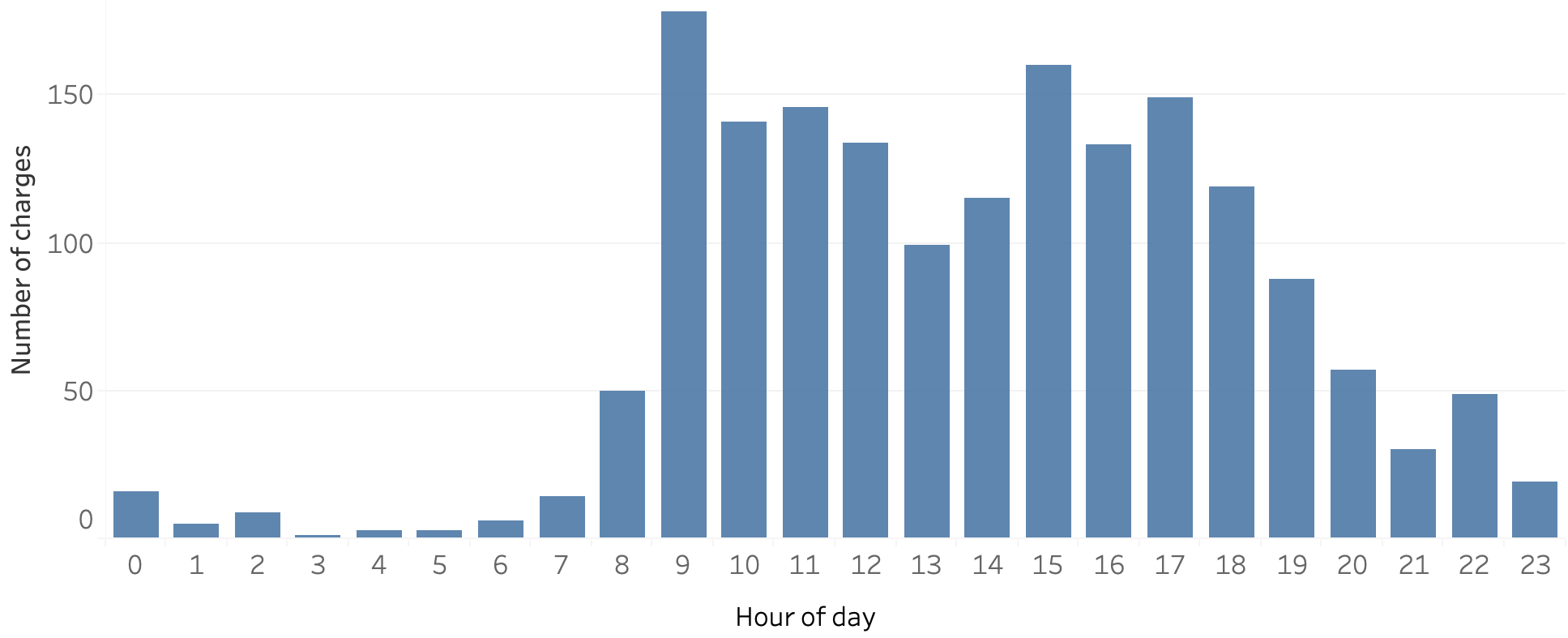}
	\caption{Distribution of charges in each hour of day.}
	\label{Fig5}
\end{figure}

\begin{figure}[htbp]
	\centering
	\includegraphics[width=14cm]{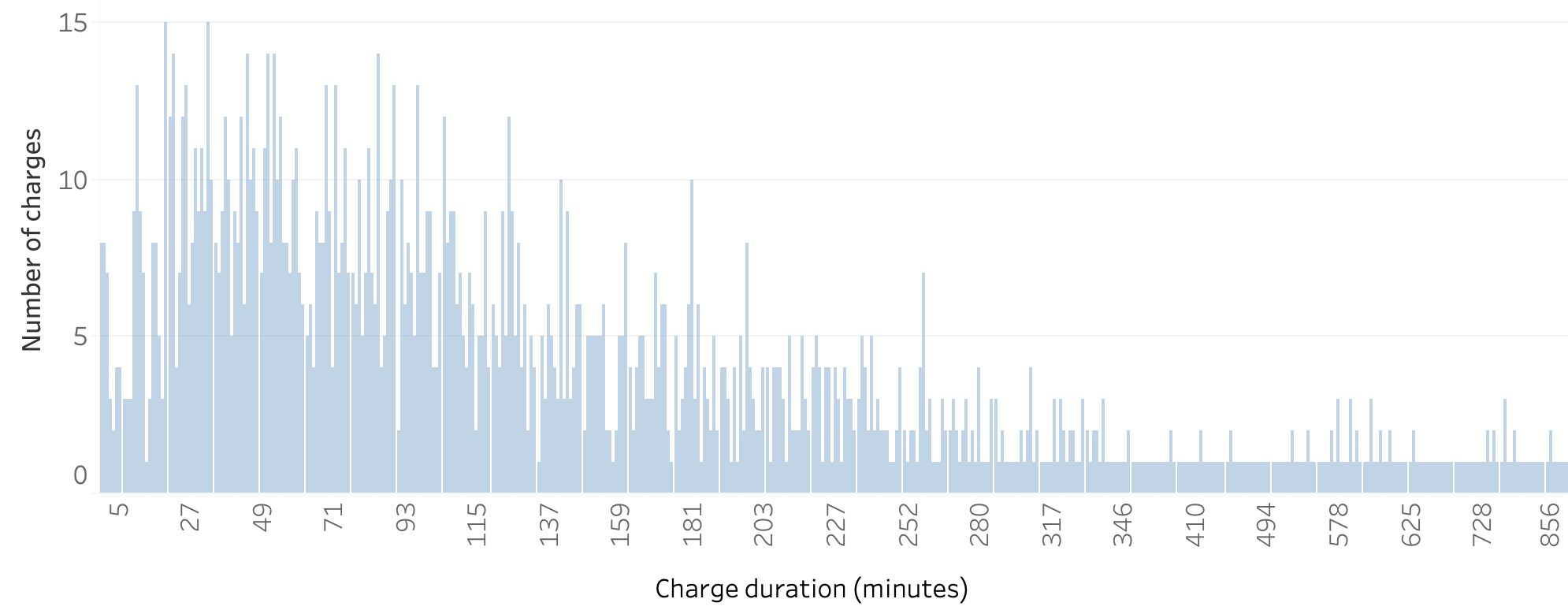}
	\caption{Distribution of charge duration in minutes.}
	\label{Fig6}
\end{figure}

\begin{figure}[htbp]
	\centering
	\includegraphics[width=10cm]{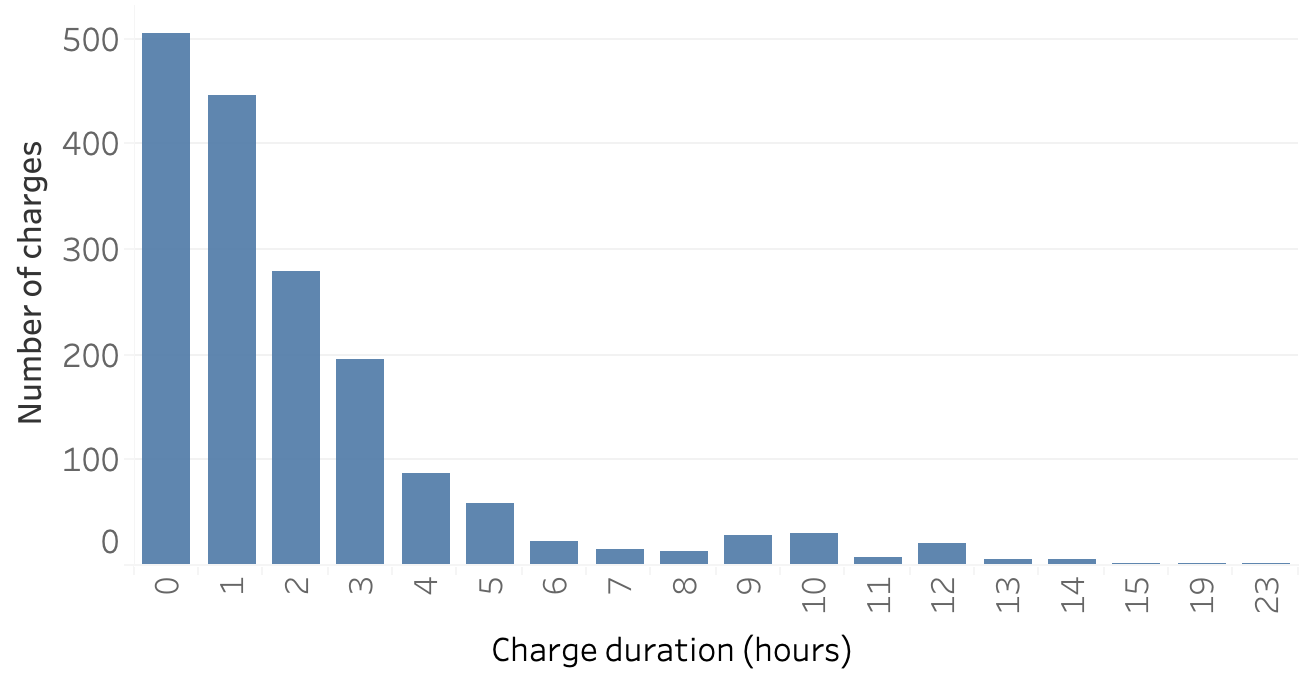}
	\caption{Distribution of charge duration in hours.}
	\label{Fig7}
\end{figure}

The data under consideration refer to 1,724 EV charges from a selected charging station. They have the following characteristics:
\begin{itemize}
\item the charges have been supplied in a period of three consecutive years;
\item the charge distributions within the different days of the week (Figure \ref{Fig4}) and the different hours of the day (Figure \ref{Fig5}) indicate a higher supply frequency in working days, with respect to Saturday and above all to Sunday, and in the hours between 9 and 18;
\item the charge duration distributions, in minutes (Figure \ref{Fig6}) and hours (Figure \ref{Fig7}), display the highest frequency of charges lasting less than one hour, with a mean distribution value around 35-40 minutes. 
\end{itemize}

\newpage
\subsection{Streaming architecture}
\begin{figure}[htbp]
	\centering
	\includegraphics[width=7cm]{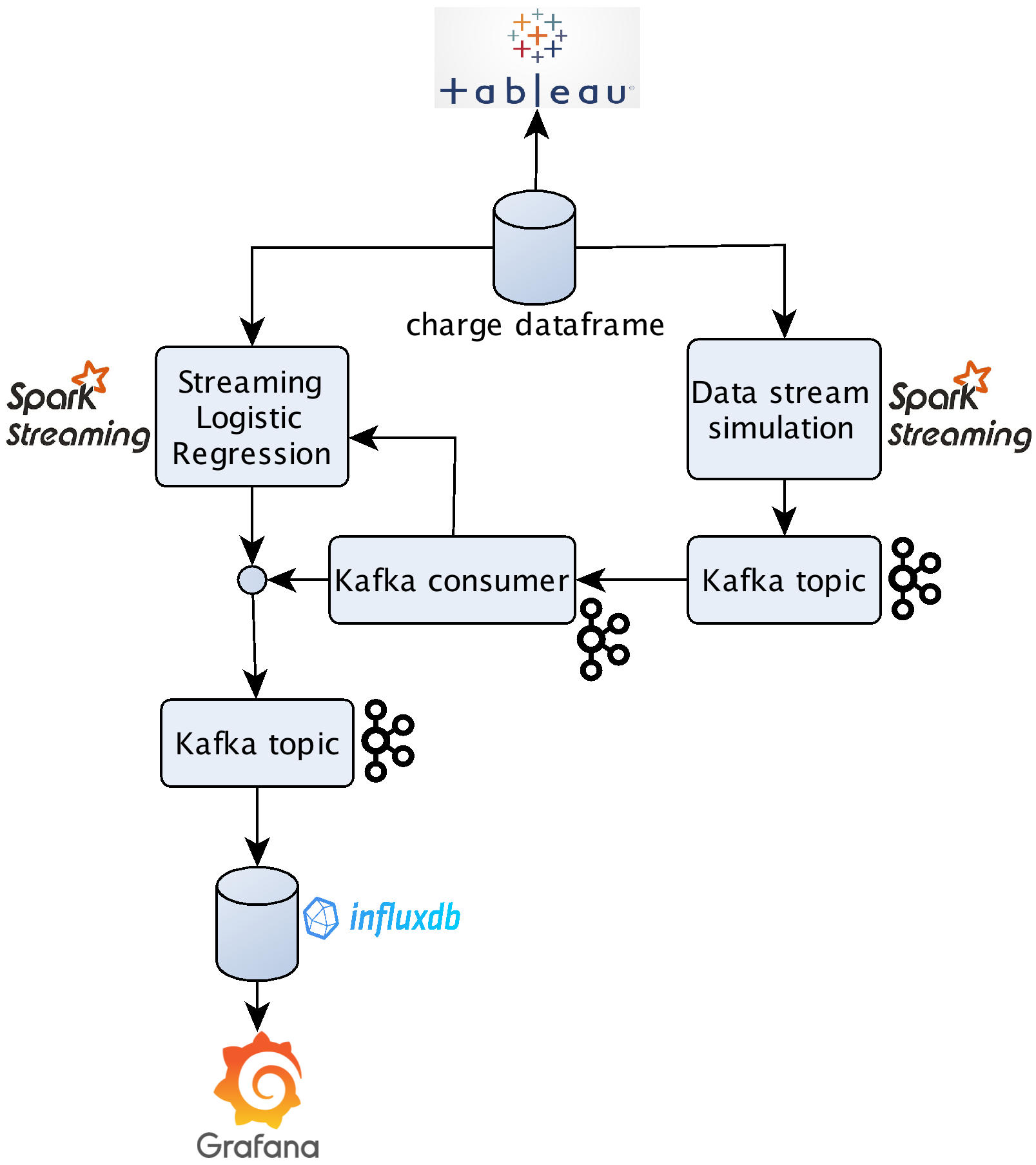}
	\caption{Developed architecture in order to create a simulated continuous data streaming, to train a \textit{StreamingLogisticRegression} model and to extract the resulting output.}
	\label{Fig8}
\end{figure}

The whole architecture has been implemented using Apache Spark and the functions from its MLlib library \cite{mllib}. In particular, the \textit{StreamingLogisticRegression} function included in the Spark MLlib library has been selected; the function is natively able to update the initialized model with the arrival of new data streams. The features chosen as model inputs are:
\begin{itemize}
\item two cyclical variables to represent hours of the day;
\item two cyclical variables to represent months of the year;
\item a categorical variable to distinguish business days from weekends;
\item a categorical variable to distinguish working days from festivities;
\item seven categorical variables to represent different days of the week.
\end{itemize}
\indent The year with the highest number of available data has been considered as a training set to initialize the Streaming Logistic Regression model. Having at disposal just static, historical data, it has been necessary to simulate continuous data streaming in order to test the architecture. This task has been realized with data from one of the two remaining years and using Apache Kafka. It is important to note that the principal aim was to demonstrate the feasibility of the streaming architecture implementation and not to select the best forecasting model.\\
\indent In Figure \ref{Fig8} the developed architecture is presented:
\begin{itemize}
\item initial dataset is imported in Tableau for preliminary analysis and visualizations;
\item test data have been selected and transformed into a continuous simulated data stream, with a time resolution of one minute; data are sent into a Kafka topic by a Kafka Producer;
\item training data have been used to initialize a \textit{StreamingLogisticRegression} model. Later, a Kafka Consumer reads the streaming data coming into the Kafka topic; this data stream has a dual functionality: on the one hand it allows an incremental update of the initialized model, on the other it is used to extract hour and date in the next 15 minutes and to provide as output the occupancy status forecast of the charging station, from the just updated model;
\item occupancy probability of the considered charging station after 15 minutes from the actual time is saved in another Kafka topic, written on InfluxDB and visualized in Grafana.
\end{itemize}
\section{Results}
\begin{figure}[htbp]
	\centering
	\includegraphics[width=15cm]{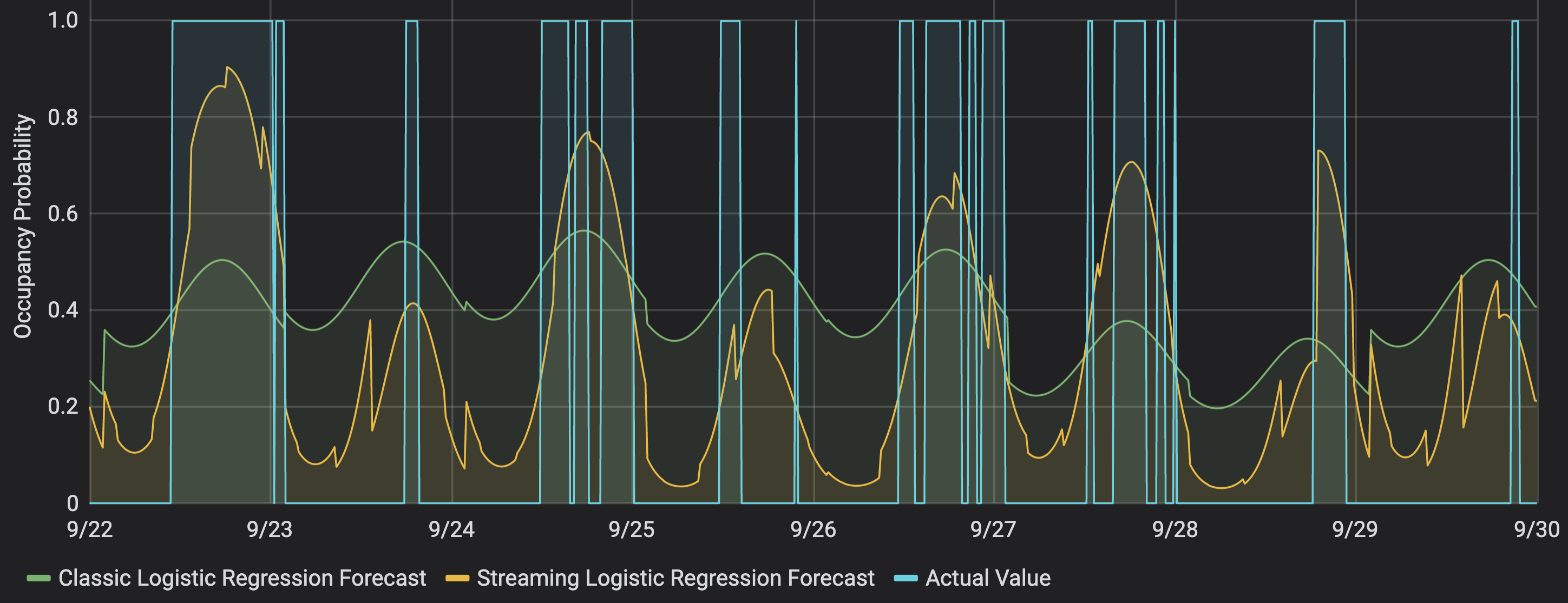}
	\caption{Occupancy probability forecasts obtained with the Streaming Logistic Regression model (in yellow) and with the classical Logistic Regression model (in green), compared to the actual occupancy status (in light-blue). The actual status is 1 if the charging station is occupied, 0 otherwise. The week between 22 and 29 September of the selected year is displayed.}
	\label{Fig9}
\end{figure}

For all 525,601 minutes in the test set, the actual charge presence or absence has been compared to the occupancy forecasts, performed 15 minutes before. Figure \ref{Fig9} displays the model predictions and the actual occupancy status for the week 22-29 September of the selected year. The considered models are the Streaming Logistic Regression model and classical Logistic Regression model, trained just using historical entries and not updated with real-time data; the actual status is 1 if the charging station is occupied, 0 otherwise. 

In order to extract the occupancy status forecast from the occupancy probability, a standard threshold of 0.5 is usually chosen: a probability higher than 0.5 indicates a charge presence, while a probability lower than 0.5 indicates a charge absence. Again the class 1 stands for a charge presence, the class 0 for a charge absence.
From a visual analysis of the results it appears that:
\begin{itemize}
\item Streaming Logistic Regression model learns from historical data a modular pattern, evident also for classical Logistic Regression model results. The occupancy probability decreases indeed during the night hours and in Saturday and Sunday, compared to the working days;
\item occupancy probabilities from the streaming model are generally lower than those from the batch model. However, if a charging station is actually occupied, the streaming forecasts display a higher increase. This increase is more evident with long charges, when occupancy probabilities reach values above 0.8; 
\item occupancy probabilities are on average lower than 0.5; therefore, the threshold to extract the corresponding class should be probably set lower than the standard of 0.5. 
\end{itemize}
The three indexes of precision, recall and F1-score allow a formalization of these results. Considering the number of false positives (FP), false negatives (FN), true positives (TP) and true negatives (TN) of the models, precision $p$ and recall $r$ are calculated as follows:
\begin{equation}
p= \frac{TP}{TP+FP} \quad \quad ; \quad \quad r =\frac{TP}{TP+FN}
\end{equation}

F1-score is the harmonic mean of $p$ and $r$:
\begin{equation}
F_1 = \frac{2}{\frac{1}{r}+\frac{1}{p}} = 2 \cdot \frac{p \cdot r}{p+r}
\end{equation}

\begin{figure}[htbp]
	\centering
	\includegraphics[width=12cm]{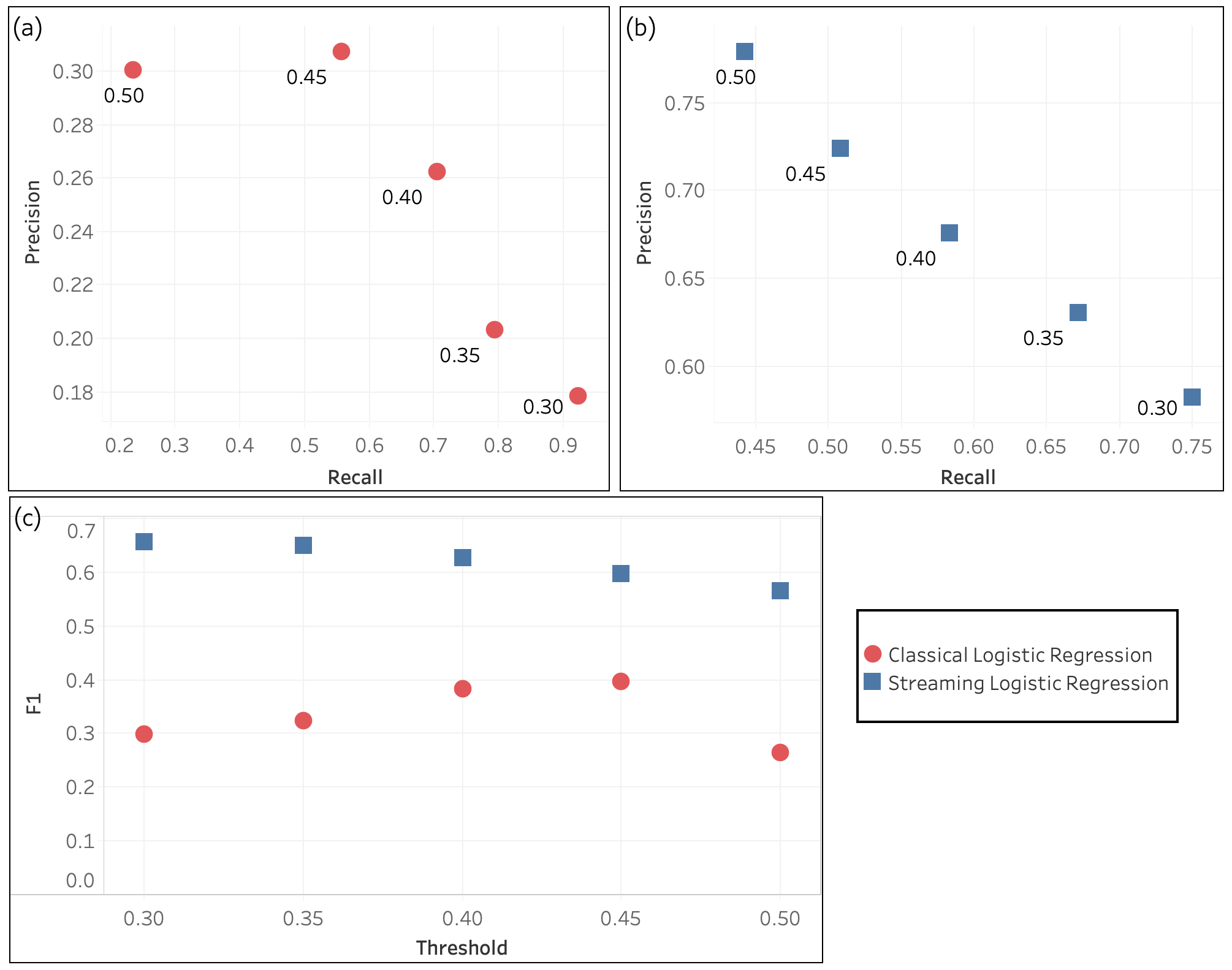}
	\caption{(a) Precision/recall curve for the classical Logistic Regression model; (b) precision/recall curve for the Streaming Logistic Regression model; (c) F1-score of the classical and Streaming Logistic Regression. Threshold values are in the range 0.30-0.50. The sample includes all 525,601 minutes in the test set.}
	\label{Fig10}
\end{figure}

Figures \ref{Fig10}a and \ref{Fig10}b show the precision/recall curve respectively for the classical and the Streaming Logistic Regressions; the sample includes all 525,601 minutes in the test set. It is evident that the streaming model has a higher precision for all the threshold values in the range 0.30-0.50, while the batch model generally presents better recalls. The combination of precision and recall in the F1-score attests more accurate predictions from the Streaming Logistic Regression, for all the set thresholds (Figure \ref{Fig10}c).
 \\
Focusing on the Streaming Logistic Regression model, the threshold value producing the best results is 0.30, while a threshold of 0.35 provides well balanced precision and recall, with values for both the indexes between 0.63 and 0.67. However, a model with a better recall than precision will be chosen in the case of need to forecast the highest number of charges, with the risk of forecasting as a charge an event that will not be confirmed as an actual charge. On the contrary, a model with a better precision than recall will be chosen when it is necessary to forecast just correct charges, with the risk of losing some charge predictions.

\section{Conclusion}
This paper presents a first model prototype to forecast the occupancy status probability of EV charging stations. The developed big data streaming architecture is based on Apache Spark, Spark Streaming and Apache Kafka. It receives streaming data from a charging station and provides as output the occupancy probability in the next 15 minutes. The selected forecasting model is the Streaming Logistic Regression, initialized using historical data and constantly updated with the arrival of real-time data streams. \\
\indent The model learns from historical data a modular pattern, with a probability decrease in the night hours and during the weekends. The real-time update of the model results in an occupancy probability increase when a charge is actually present. Therefore the streaming model provides better predictions than the batch model.
The occupancy status retrieval has been done by fixing different threshold values: if the occupancy probability is higher than the set threshold the prediction states the charging station occupancy, otherwise it states the charging station availability. A threshold of 0.35 allows to seek a balance between precision and recall indexes, resulting in the range 0.63-0.67 in this case. \\
\indent The results highlight the necessity of a further optimization of Logistic Regression parameters, such as the regularization parameters, the streaming time window, the selected features and the gradient descent step. As regards the choice of the classification model, just the Logistic Regression model has been tested so far, but it is necessary to investigate which model is the best appropriate for the specific use case; other examples could be the Decision Tree Classifier, the Random Forest Classifier and the Gradient-Boosted Tree Classifier \cite{classification}. \\ 
\indent Moreover, a web or mobile application could be developed to display on a map the resulting occupancy probabilities in real-time for all the available charging stations. This application will provide a tool of easier and more immediate use, supporting an EV driver in the choice of the most probable free charging station, close to his destination. \\
\indent Finally, it could be useful to investigate the weights in the mix between batch and real-time data and understand how to calibrate this mix. This ability to keep historical and actual situations both into account can be considered indeed as the main strength of the proposed architecture and will become increasingly important for forecasting models related to various fields in a fast and continuously changing world.

\bibliographystyle{IEEEtranN}
\bibliography{references}  
\end{document}